\documentclass[letter, 10pt, conference]{ieeeconf}      % Use this line for a4 paper
\IEEEoverridecommandlockouts % 
\usepackage{times}
\usepackage{graphicx}\graphicspath{{img/}}
\usepackage{epstopdf}
\usepackage{algorithm}
\usepackage{algorithmic}
\usepackage{multirow} 
\usepackage{multicol}
\usepackage{array}
\usepackage{mdwmath}
\usepackage{mdwtab}
\usepackage{rotating}
\usepackage{amsmath}
\usepackage{amssymb}
\usepackage{verbatim}
\usepackage{overpic}
\usepackage[labelformat=simple]{subcaption}

\usepackage[utf8]{inputenc}
\usepackage[T1]{fontenc}
\usepackage[export]{adjustbox}
\usepackage[normalem]{ulem}
\usepackage{siunitx} % for formatting units
\usepackage{chngpage}
\usepackage{xspace}

\usepackage[acronym]{glossaries}%
\glsdisablehyper % switch off hyperlinks for abbreviations
\makeglossaries%

\newacronym{LfD}{LfD}{learning from demonstration}\providecommand{\LfD}{\gls{LfD}}%
\newacronym{MT}{MT}{machine teaching}\providecommand{\MT}{\gls{MT}}%
\newacronym{TD}{TD}{teaching dimension}%
\newacronym{nmse}{NMSE}{normalised mean squared error}
\newacronym{ols} {OLS} {ordinary least squares}
\newacronym{LiP}{LiP}{linear-in-the-parameters}%
\newacronym{SVD}{SVD}{singular value decomposition}%

\usepackage{amssymb}%
\usepackage{mathtools}%
\usepackage{amsmath}%
\usepackage{gensymb}%
\usepackage{bm}%

\mathchardef\mhyphen="2D   % define "math hyphen"

\providecommand{\R}     {\mathbb{R}}          % real numbers
\providecommand{\T}     {\top}                % transpose
\providecommand{\I}     {\bm{I}}          % identity matrix
\providecommand{\estimated} [1]{\tilde{#1}}

\providecommand{\optimal}   [1]{\bar{#1}}

\DeclareMathOperator{\vect}{vec}% vectorisation operator

\providecommand{\nd}      {n}                              % index for data points
\providecommand{\nx}     {p}                  % index for elements of input/state vector
\providecommand{\dimx}   {\mathcal{\MakeUppercase{\nx}}} % input/state dimensionality
\providecommand{\dimy}   {\mathcal{\MakeUppercase{\ny}}} % output/observation dimensionality

\providecommand{\q}      {q}                  % joint angles
\providecommand{\bq}     {\bm{\q}}        % (bold) joint angles
\providecommand{\br}     {\bm{r}}         % (bold) end-effector positions
\providecommand{\bJ}     {\bm{J}}         % Jacobian
\providecommand{\bqdot}  {\dot{\bq}}          % (bold) joint velocities
\providecommand{\brdot}  {\dot{\br}}          % (bold) end-effector velocities
\providecommand{\bqddot} {\ddot{\bq}}         % (bold) joint accelerations
\providecommand{\u}      {u}                  % input/command

\providecommand{\bu}     {\bm{\u}}        % (bold) input/command
\providecommand{\nu}     {q}                  % index for elements of u
\providecommand{\dimu}   {\mathcal{\MakeUppercase{\nu}}} % action dimensionality
\providecommand{\bpi}    {\bm{\pi}}   % (bold) control policy
\providecommand{\bphi}  {\bm{\phi}}               % feature vector
\providecommand{\bPhi}  {\bm{\Phi}}               % 
\providecommand{\nphi}  {s}                               % index for feature vector elements
\providecommand{\dimphi}{\mathcal{\MakeUppercase{\nphi}}} % feature dimensionality
\providecommand{\numD} {\mathcal{N}} % number of data points

\providecommand{\A}     {\mathcal{A}} 
 
\renewcommand{\dimy}   {\mathcal{R}} % output/observation dimensionality
\renewcommand{\dimu}{\dimy}%

\providecommand{\model}{\bm{\theta}}  % model
\providecommand{\Model}{\bm{\Theta}}  % model
\providecommand{\target}{{\model}}  % target model
\providecommand{\Target}{{\Model}}  % target model

\providecommand{\learnt}{\estimated{\bm{\theta}}}  % learnt model
\providecommand{\learntaction}{\estimated{\action}}  % learnt model

\providecommand{\dataset}{\bm{\mathcal{D}}}          % data set
\providecommand{\optimaldataset}{\optimal{\dataset}} % optimal data set
\providecommand{\textltwonorm}{$\mathcal{\ell}^2$-norm}

\providecommand{\searchspaceD}{\bm{\mathfrak{D}}} % search space of data set
\renewcommand{\u}{u} % torque/action
\providecommand{\action}{\bm{u}}  % torque/action

\providecommand{\state}{\bm{x}}         % (bold) state 
\providecommand{\actions}{\mathbf{u}}  % vector containing action samples - this is not the same as the action vector for multi-dimensional actions
\providecommand{\optimalactions}{\optimal{\actions}}  % torque/action matrix

\providecommand{\featuref}  {\bm{\mathcal{\phi}}}    % basis function

\providecommand{\triskf}  {\mathcal{\rho}}    % teaching risk function
\providecommand{\terrorf}  {\mathcal{\varepsilon}}    % teaching effort function
\providecommand{\terrorfmax}{\terrorf_{max}}          % maximum permissible teaching effort

\providecommand{\ltwonorm}  {\mathnormal{E_{\ell_2}}}

\providecommand{\length}      {l}   %length
\providecommand{\mass}      {m}   %mass
\providecommand{\gravity}      {g}   %gravity

\providecommand{\q}      {q}           % joint angle
\providecommand{\qi}     {\q_1}        % first joint angle
\providecommand{\qii}    {\q_2}        % second joint angle
\providecommand{\qdoti}  {\dot{\q}_1}  % first joint angular velocity
\providecommand{\qdotii} {\dot{\q}_2}  % second joint angular velocity
\providecommand{\bM}     {\bm{M}}      % mass matrix
\providecommand{\bC}     {\bm{C}}      % Coriolis matrix
\providecommand{\bg}     {\bm{g}}      % gravity vector

\renewcommand  {\r}      {r}           % d 
\renewcommand  {\xi}      {\r_1}           % first state variable
\providecommand{\xii}     {\dot{\r}_1}           % second state variable

\providecommand  {\yi}      {\r_2}           % first state variable
\providecommand{\yii}     {\dot{\r}_2}           % second state variable

\providecommand{\fx}     {f_1}           % first force variable
\providecommand{\fy}     {f_2}           % second force variable

\providecommand{\udx}{\Delta r_{1}}           % first physical action variable
\providecommand{\udy}{\Delta r_{2}}           % second physical action variable

\providecommand{\rd}{\optimal{\r}}%
\providecommand{\targetx}{\rd_1}%
\providecommand{\targety}{\rd_2}%

\providecommand{\nsubjects}{n}  % number of subjects
\providecommand{\screenx}{518}
\providecommand{\screeny}{325}

\DeclareUnicodeCharacter{0308}{??}

\usepackage{xspace}

\providecommand{\ie}{\textit{i.e.,}~} %
\providecommand*{\sref}[1]{\S\ref{s:#1}}            % section

\providecommand{\figurename}{Fig.}
\providecommand*{\fref}[1]{\figurename~\ref{f:#1}}  % figure
\providecommand*{\eref}[1]{(\ref{e:#1})}            % equation

\let\labelindent\relax %<-this prevents a conflict with IEEE \labelindent when using enumitem
\usepackage[inline]{enumitem} % use enumitem package to get inline lists, set the list format
\setlist{nolistsep}
\providecommand{\il}[1]{\begin{enumerate*}[label=(\roman*)]#1\end{enumerate*}} % inline list (for use in main text)
\usepackage[normalem]{ulem}                                        % for striking text out
\usepackage{marginnote}
\usepackage[textwidth=\marginparwidth,colorinlistoftodos]{todonotes}

\colorlet{jb}{red}
\colorlet{mh}{red}
 
\providecommand  {\colorsout}[1]{\bgroup\markoverwith{\textcolor{#1}{\rule[0.5ex]{2pt}{0.4pt}}}\ULon} % define coloured strike out

\colorlet{ma}{blue}

\makeatletter\newcommand{\manuallabel}[2]{\def\@currentlabel{#2}\label{#1}}\makeatother

\usepackage[
  backend=bibtex,
  style=ieee,
  url=false,
  isbn=false,
  doi=true,
  eprint=false
]{biblatex}

\usepackage{gensymb}
\usepackage[acronym]{glossaries}
\usepackage{pgfplots}
\usepackage{filecontents}
\usepgfplotslibrary{fillbetween}
\usetikzlibrary{arrows.meta}
\usepackage{float}

\usepackage{tikz}
\usepgfplotslibrary{statistics}
\usepackage{siunitx}

\title{\LARGE \bf
Using Machine Teaching to Boost Novices' Robot Teaching Skill}

\author{Yuqing Zhu$^{1}$, Endong Sun and Matthew Howard% <-this % stops a space
\thanks{*This work was supported by King's College London and the China Scholarship Council.}% <-this % stops a space
\thanks{$^{1}$Yuqing Zhu ({\tt\small yuqing.zhu@kcl.ac.uk}), Endong Sun and Matthew Howard are with the Centre for Robotics Research, Department of Engineering, King's College London, UK.}%
}

\begin{document}%

\maketitle
\thispagestyle{plain}%
\pagestyle{plain}%

\begin{abstract}
Recent evidence has shown that, contrary to expectations, it is difficult for users, especially novices, to teach robots tasks through \LfD. This paper introduces a framework that leverages \emph{\acrlong{MT}} algorithms to train novices to become better teachers of robots, and verifies whether such teaching ability is \il{\item \emph{retained} beyond the period of training and \item \emph{generalises} such that novices teach robots more effectively, even for skills for which training has not been received}. A between-subjects study is reported, in which novice teachers are asked to teach simple motor skills to a robot. The results demonstrate that subjects that receive training show average 78.83\% improvement in teaching ability (as measured by accuracy of the skill learnt by the robot), and average 63.69\% improvement in the teaching of new skills not included as part of the training.
\end{abstract}\glsresetall%

\section{Introduction}
\label{s:intro}
Robot \LfD\ is a technology that enables robots to learn tasks by observing and imitating human actions, eliminating the need for explicit programming. This approach greatly expands accessibility, allowing non-experts to train robots, and makes their deployment much simpler in a wide range of fields \cite{bechar2017agricultural,bojarski2016end}. \LfD\ stands out for its ability to facilitate the learning of complex tasks efficiently, becoming a crucial direction in the advancement of modern robotics \cite{breazeal2016social}.

The human teacher's role in \LfD\ is critical, as the quality of their demonstrations directly influences the robot's ability to learn and generalise tasks. Inexperienced teachers may provide ambiguous demonstrations, leading to sub-optimal robot behaviours \cite{atkeson1997robot}. To overcome this, there is growing interest in the concept of \emph{training human teachers} to deliver high-quality demonstrations, improving both \LfD\ outcomes and the teachers' capabilities across tasks. For example, \cite{Sena2020} focuses on identifying key teaching behaviours and providing feedback based on optimal demonstration sets. However, reliance on pre-selected demonstrations limits generalisation to new skills. Similarly, \cite{aoyama2021training} proposed adaptive feedback based on numerical analysis of demonstrations to train human teachers. However, this approach has only been tested on a basic torque-controlled pendulum system, which may not generalise well to more complex systems and dynamic motor skills. Thus, current methods are either limited in scalability or constrained by the simplicity of their application. This highlights a significant research gap: the need for more scalable and generalisable approaches to training human teachers, ensuring \LfD\ systems can adapt to diverse and evolving robotic applications.

\begin{figure}[t]
    \centering
    \includegraphics[width=\linewidth]{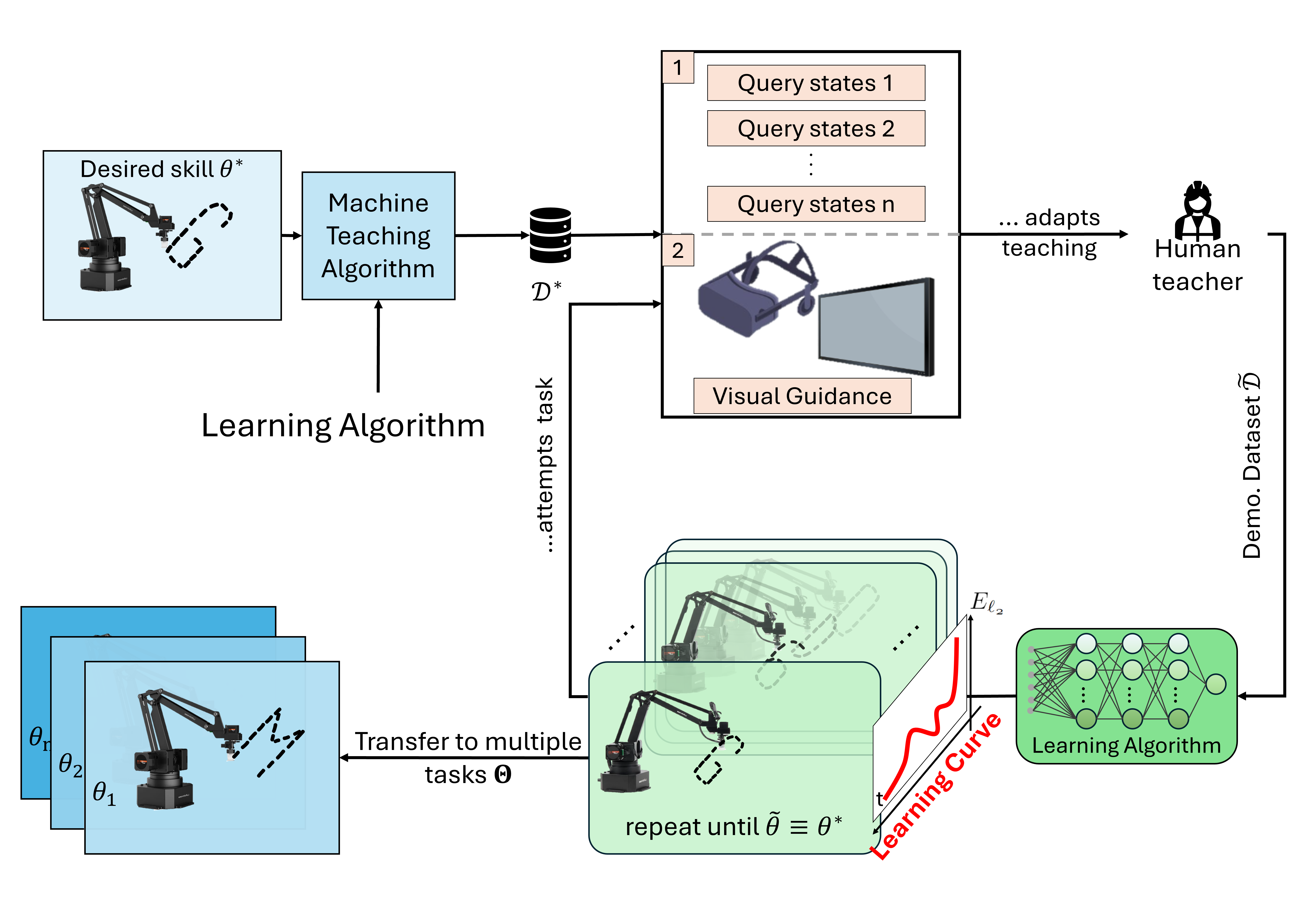}
    \caption{Framework overview. The proposed approach allows human teachers to be trained to teach robot dynamic motor skills using \acrlong{MT}.}
    \label{f:overview_flowchart}
\end{figure}
Building on these foundations, this paper introduces a novel training framework to enhance teaching quality within \LfD\ by integrating \MT\ into the process \cite{Zhu2015}, as shown in \fref{overview_flowchart}. The goal is not only to improve teaching \emph{quality} but also to streamline the process by reducing the time and number of demonstrations required. This begins with considering both student (\ie robot) \emph{and} teacher perspectives within the \LfD\ process, in order to design a system of guidance for training novice teachers across different tasks. %score derived from \MT\ algorithm is then introduced, providing a general framework for \LfD\ challenges that considers both the robot student's learning performance and the teaching efforts. 
An empirical analysis of the proposed framework is conducted through experiments in which subjects teach a robotic arm simple motor skills. %with a set of $\nsubjects=\nsubjectsa$ subjects to validate the effectiveness of the proposed feedback mechanism. Multi-phase experiments 
The findings demonstrate how the framework \il{\item enhances the quality of teaching of novice participants, \item this is retained past the period of training, and \item this generalises to the teaching of motor skills not addressed in the training}. %Moreover, these experiments show how teaching skill obtained through training transfers across tasks. To summarise, this work makes the following contributions:
%\begin{itemize}
%    \item A general framework for feedback mechanisms using an \MT\ algorithm is proposed, enabling the integration of new skills without human expert intervention and reducing reliance on predefined demonstration sets.
%    \item The feedback mechanism is tested in teaching motor skills to robotic manipulators, with experiments in both simulation and real-world settings, demonstrating its effectiveness in complex and practical scenarios.
%    \item The potential for skill transferability is investigated, showing that the framework can facilitate learning across different tasks.
%\end{itemize}
%  
The effect is seen to be statistically significant, with average 78.04\% improvement in teaching performance for the subjects who undergo training (as measured by the accuracy with which the skill is learnt). %Furthermore, the feedback system improves skill transferability, leading to a 70\% improvement for the target group, while the control group experienced a 10\% degradation.

\section{Problem Definition}%
\label{s:problem_definition}%
In the problems considered in this paper, motor skills are represented as a control policy in the form
% + ref
\begin{equation}\label{e:SkillModel} 
    \action = \bpi(\state,\model)
\end{equation}
where $\bpi(\cdot)$ maps the robot's state, $\state \in \R^\dimx$, to the corresponding action, $\action \in \R^\dimu$. The variable $\model \in \R^\dimphi$ denotes the skill parameters. 

Skills are communicated to the robot through teaching, \ie by providing a set of demonstration data $\dataset$ consisting of a sequence of state-action tuples $\dataset = \{(\state_1,\action_1) ...,(\state_\numD,\action_\numD)\} \in (\R^{\dimx\times\numD}, \R^{\dimu\times\numD})$. With this, the robot learner learns the skill parameters with a learning algorithm
\begin{equation}
\learnt = \A (\dataset) \label{e:A(D)}
\end{equation}
leading to the robot learning the skill
\begin{equation}
	\learntaction = \bpi(\state,\learnt). \label{e:estimatedpi(x)}
\end{equation}
Note that, the success of learning depends heavily on the \emph{quality and quantity of the demonstration data}, something that can not easily be assured when the data is provided by novice human teachers. %If such standards are missed, it can lead to poor learning outcomes for the robot. 

One approach to deriving high-quality data is to apply the principles of \acrlong{MT}. %Given these challenges, it is crucial to employ strategies that refine and optimise the training data set to maximised the efficacy and efficiency of learning. This is precisely the focus of the \MT\ algorithm. \MT\ can be considered as the \emph{inverse} of conventional machine learning. 
In \MT, the aim is to find the optimal data that will lead a given learner, to learn a target model $\target$. One formulation of this is as
%Generally, \MT\ can be formulated as 
a bi-level optimisation problem \cite{Zhu2018}
\begin{align}
	\optimaldataset&=\arg\min_{\dataset \in \searchspaceD} 
\triskf (\learnt, \target) \label{e:MTBilevel1} \\ 
	\mathrm{s.t.}\quad \learnt&= \A (\dataset) %\label{e:MTBilevel2} \\
	%&
 \quad\mathrm{and}\quad \terrorf (\dataset)\le \terrorfmax. \label{e:MTBilevel3}
\end{align} 
Here, %\eref{MTBilevel1} and \eref{MTBilevel3} define the teacher's problem. The teacher's 
%the aim is to find a data set $\dataset$ from the space of all possible data sets $\searchspaceD$ that minimises 
the teaching risk $\triskf(\cdot)$ measures the accuracy with which the target model is learnt using a data set from the space of all possible data sets $\searchspaceD$ and $\terrorf(\cdot)$ represents the teaching effort (which, in this formulation, is constrained to up to a maximum of $\terrorfmax$). %Note that, the optimisation requires knowledge of This must be solved subject to the learner deriving its learnt model through \eref{MTBilevel2}.

The solution of \eref{MTBilevel1}-\eref{MTBilevel3} requires \il{\item the target model $\target$ is explicitly known and \item the optimisation to be tractable}. In many cases, this cannot be assured, %\il{\item the target model is only known \emph{implicitly} and \item the \MT\ problem is intractable,} 
especially when considering complex robotic motor skills. %or the learning algorithm does not have a closed-form solution. 
However, this is where human teachers' (often implicit) knowledge of a broad range of motor skills may be exploited: this study explores if and how \emph{the optimality benefits of \MT\ can be combined with human versatility}, by shaping their teaching strategies toward those deemed optimal according to \MT\ thorugh training. It is anticipated that this can lead to a significant improvement in the overall quality of teaching, for example, enabling humans to transfer optimal teaching of one motor skill to another, including those for which the \MT\ problem cannot be directly solved.

\begin{enumerate}[label=$h_\arabic*$]
    % \item \label{h0}%\textbf{H1:} Teaching quality does not improve when humans are instructed by guidance derived from the principles of \MT.
    \item \label{h1}%\textbf{H1:} 
    Teaching quality improves when humans are instructed by guidance derived from the principles of \MT.
    \item \label{h2}%\textbf{H2:} 
    After guidance is removed, the improved teaching ability is retained.
    \item \label{h3}%\textbf{H3:} 
    The improved teaching ability learnt in one skill can be transferred to another skill.
    % \item The quality of the demonstration provided by the human teacher improves when they are instructed by guidance in the simulator.
    % \item The quality of the demonstration provided by the human teacher improves when they are instructed by guidance with a physical robot.
    %\item After guidance is removed, the improved teaching ability of human teachers on skills is retained.
    %\item The teaching abilities learnt by human teachers in one skill can be transferred to another skill.
\end{enumerate}

The next section describes the materials and methods used to test these hypotheses.
% \section{Methodology}
\section{Materials \& Methods}
\label{s:methodology}%
\subsection{\MT\ Formulation}
The core to the proposed framework is to derive guidance from \MT\ as a training signal for novice human teachers. For this, the following modelling assumptions and design choices are made.
\subsubsection{Modelling the Learner}
%In \LfD, a key challenge is enabling the robot to generalise from a small set of noisy demonstrations. These demonstrations, provided by human teachers, often contain variability and limited coverage of the task space. As such, the learner model must balance simplicity and robustness to handle this imperfect data while efficiently extracting useful patterns. Given these constraints, we selected a linear control policy:
To balance the demands of simplicity, robustness and scalability, in this paper, all robot learners use control policies of the form
\begin{equation}\label{e:LiPSkillModel} 
    \u = \target^\T\featuref(\state)
\end{equation}
where $\featuref(\state) \in \R^\dimphi$ is a mapping from the state to a suitable feature space. Note that, depending on the choice of features, this model may represent policies of arbitrary complexity, and enables the inclusion of robot-specific features (such as kinematic mappings, see \sref{evaluation}). \footnote{For simplicity, the following describes learning in the case that $\dimu=1$, this may be trivially extended to $\dimu>1$ by replacing $\target$ with the appropriate skill parameter matrix $\Target$.} %This model is both computationally efficient and flexible enough to capture the essential structure of the data, making it well-suited to real-world robotic applications where data sparsity and noise are unavoidable.
%
%More complex models, such as deep neural networks or Gaussian processes, may seem appealing due to their ability to model complex dynamics. However, they come with significant drawbacks: they require large amounts of data to avoid overfitting and introduce unnecessary computational overhead. For our problem, where demonstrations are limited and generalisation is paramount, such models are not appropriate. Instead, a linear policy offers a clearer, more interpretable structure, ensuring the robot learns efficiently without the risk of overfitting to a small dataset.

The parameter estimate $\learnt$ is found through ridge regression %To estimate the parameters of this control policy, ridge regression is applied with the objective 
% \begin{equation} \label{e:LipLoss}
% \min\limits_{\learnt \in \R^\dimphi} \sum_{i=1}^{\numD} \frac{1}{2}(\learnt^\T \bphi_i - \action_i)^2 + \frac{\lambda}{2} |\learnt|^2 
% \end{equation} 
%
%\begin{equation} \label{e:LipLoss}
%\min\limits_{\learnt \in \R^{\dimphi\times2}} \sum_{i=1}^{\numD} \frac{1}{2}(\learnt^\T \bPhi_i  - \bm{\action_i})^2 + \frac{\lambda}{2} \| \learnt \|^2
%\end{equation} 
%where $\lambda$ is a regularisation parameter.This method is particularly suited to our task because it prevents overfitting through regularisation, making the model more robust in handling noisy and small datasets. Its solution:
for which the closed-form solution is
\begin{equation} \label{e:LipSolve} 
    \learnt=(\bPhi^\T\bPhi + \lambda \I)^{-1}\bPhi^\T \actions
\end{equation}
%
%
%
%\begin{equation} \label{e:LipLoss}
%\min\limits_{\learnt \in \R^{\dimphi\times2}} \sum_{i=1}^{\numD} \frac{1}{2}(\boldsymbol{\Theta}^\T \bPhi_i  - \bm{\action_i})^2 + \frac{\lambda}{2} \| \boldsymbol{\Theta} \|^2
%\end{equation} 
%\begin{equation} \label{e:LipSolve} 
%    \boldsymbol{\Theta} =(\bPhi^\T\bPhi + \lambda \I)^{-1}\bPhi^\T \actions.
%\end{equation}
%
%
%provides a simple, closed-form approach that is computationally efficient and reliable.
% .
where $\actions=(\u_1,...,\u_\numD)$, the $\nd$th column of $\bPhi\in\R^{\dimphi\times\numD}$ contains $\bphi(\state_\nd)$ and $\lambda$ is a regularisation parameter.
Assuming the robot learns through \eref{LipSolve}, the specification of the \MT\ problem requires that appropriate teaching risk and effort functions are selected. 
\subsubsection{Teaching Effort}
\label{s:teachingeffort}
%Several metrics could be used to quantify teaching effort, such as the time taken in teaching, or number of interactions required with the robot. 
In this study, the \emph{number of demonstrations given by the teacher} as a parsimonious, platform-independent measure of effort \ie %The strength of using \TD\ lies in its ability to streamline the teaching process. By focusing on the smallest necessary data points, it allows the human teacher to provide concise demonstrations.
\begin{equation} \label{e:TeachingBudget}
    \terrorf(\dataset) = \numD.
\end{equation}
Note that, in \MT\ formulation \eref{MTBilevel1}-\eref{MTBilevel3} imposes an upper bound on the teaching effort. Considering \eref{LipSolve},  the minimal number of data points required for effective learning can in fact be quantified as the minimum number of data points required to span the feature space \cite{Liu2016}, so $\terrorfmax=\dimphi$ % Therefore %, \ie
%\begin{equation} \label{e:TeachingBudget}
%    \terrorfmax = \dimphi
%\end{equation}
is set as the upper bound in teaching effort.

\subsubsection{Teaching Risk}
%One way to measure teaching risk is to compute the discrepancy between $\learnt$ and $\target$, which equivalently captures the distance between the learned and the target policies. While other approaches, such as probabilistic models or margin-based metrics, could be considered, we chose this $L_2$-norm discrepancy due to its computational simplicity and suitablity for real-time \LfD applications, where quick feedback on teaching quality is crucial. Furthremore, the metric aligns well with the linear learner model used in this study, allowing for efficient integratino into the overall framework.
In this study, the \textltwonorm\ between the learnt and target parameters is used as the teaching risk, \ie
%Mathematically, this risk is expressed as:
\begin{equation}
\label{e:RiskFunction}
    \triskf (\learnt, \target) = \|\learnt - \target\|_2.
\end{equation}
This risk function provides a direct measure as to whether teaching has resulted in learnt parameters that are close to the target, making it suitable for controlled experiments, such as those presented here, where $\target$ is explicitly known.

\subsection{\MT-based Training}
The training framework used in this study uses direct guidance on the quality of demonstrations to uses to shape their teaching behaviour. For this, guidance is derived from an analysis on how demonstrations affect the learnt parameter based on \MT. 

\subsubsection{Selection of States}%
\label{s:selection_of_states}%
To scaffold learning, states are pre-selected for the trainee, rather than having them provide both states and actions while demonstrating. This approach is supported by recent work \cite{Sena2020}, which shows that humans often struggle to select states that effectively test a robot's ability to generalise. Additionally, it has been shown that for the learning algorithm \eref{LipSolve}, the optimal choice of states is non-unique \cite{Zhu2015}, a fact that may cause confusion for non-technical trainee teachers. %highlights redundancy in state selection for linear problems, where not all states contribute equally to learning. By controlling the state selection, both the efficiency of learning and the robustness of the system are maximised.

In the experiments reported in this study, states are selected through an pseudo-random process, in which states are iteratively added to a set of so-called \emph{query states} $\{\state_1,...,\state_\numD\}$, with the only constraint that each state appended to the set should not cause $\bPhi$ to become ill-conditioned.

\subsubsection{Selection of Actions}%
Given the query states, users are then required to select the corresponding actions $\{\u_1,...,\u_\numD\}$ to complete the data set and demonstrate the skill. For this, the \MT\ risk function is used to guide their choice, as follows.

Substituting \eref{LipSolve} into \eref{RiskFunction}, yields
\begin{equation}
\triskf (\learnt, \target) = \left\|\left(\bPhi^\top \bPhi + \lambda \I\right)^{-1} \bPhi^\top\actions - \target\right\|_2.
\end{equation}
Since $\target$ is independent of $\bPhi$ and $\actions$, %it can be assumed for calculating the teaching risk that $\target$ shares the same $\bPhi$ as $\learnt$ but corresponds to different actions. Thus, the risk function simplifies to:
this can be simplified to
\begin{equation}
\triskf (\learnt, \target) = \left\|\left(\bPhi^\top \bPhi + \lambda \I\right)^{-1} \bPhi^\top (\actions - \optimalactions)\right\|_2.
\end{equation}
where $\optimalactions=(\optimal{\u}_1,...,\optimal{\u}_\numD)$ are the \emph{optimal} action demonstrations (\ie those that would lead to $\learnt=\target$ when learnt via \eref{LipSolve}).

Using the submultiplicative property of matrix norms \cite{horn2012matrix}, this can be written as
\begin{equation}
\triskf (\learnt, \target) \leq \underbrace{\left\|\left(\bPhi^\top \bPhi + \lambda \I\right)^{-1} \bPhi^\top\right\|_2}_{\triskf_1} \underbrace{\|\actions - \optimalactions\|_2}_{\triskf_2}.
\end{equation}
Here, %$\triskf_1$ reflects the sensitivity of $\learnt$ to $\bPhi$, while 
$\triskf_2$ captures %the discrepancy between actions. %To optimise this, both terms must be minimised simultaneously.
%\subsubsection{Minimising $\triskf_1$}
%This term is related to how the matrix \textltwonorm\ behaves when transforming vectors. This \textltwonorm\ corresponds to the maximum singular value after \SVD\ and reveals how well the matrix handles numerical stability. Simply, a smaller \textltwonorm\ indicates a more accurate and stable result when solving for $\learnt$. Thus, minimising $\triskf_1$ improves the numerical stability of the system. To achieve this, it is essential to ensure that $\bPhi$ is well-conditioned and not close to singular, which can be done by avoiding highly correlated features and ensuring a sufficient number of data points relative to the feature dimensionality.
%
% The \textltwonorm\ of the first component corresponds to the maximum singular value after \SVD. The smaller this maximum singular value, the smaller the \textltwonorm. Additionally, a smaller maximum singular value implies that the matrix $\left(\bPhi^\top \bPhi + \lambda \I\right)^{-1} \bPhi^\top$ exerts a weaker magnifying effect on the vector.
% %
% In other words, this matrix exhibits a reduced scaling effect in Euclidean space, indicating that the difference between the two vectors remains relatively small after the transformation, thereby facilitating the minimisation.
%
%\subsubsection{Minimising $\triskf_2$}
%The latter effectively The \textltwonorm\ of the second component, $\triskf_2$, represents 
the difference between the actions provided during \LfD\ and the desired actions. %The larger the discrepancy, the further the demonstrated actions deviate from the target actions, indicating that the robot is learning from suboptimal data. 
This provides a clear, actionable, and quantifiable measure of demonstration quality, allowing for immediate evaluation of how the demonstrations can be improved, and therefore forms the basis of the training guidance.

\subsection{Guidance Design}
The key considerations used to develop the proposed guidance system are that the feedback must \il{\item convey all critical pieces of information, such as how much efforts need to spend, the direction and magnitude of the actions, \item be intuitive, enabling the novice teacher to interpret without requiring extensive technical knowledge and \item  be timed such that it is provided at the right moment (too early may disrupt the demonstration, while too late risks the teacher forgetting details).} 
%
%In the study, many approaches such as verbal feedback, overall scoring and graphical summaries were considered but found unsuitable for this context. Verbal feedback, while flexible, often lacks the precision needed to communicate specific action details. Providing an overall score, though useful for assessing performance, does not offer actionable insights into how to improve specific aspects of the demonstration. Graphical summaries, such as heatmaps, can give a more detailed overview but often overwhelm the user with too much information at once.

 % In particular, feedback consists of the display of arrows on a display screen, representing \il{\item the correct action $\optimalactions$ and \item the action provided by the user $\bu$}. 

Based on these considerations, post-demonstration visual feedback is used as the clearest and most informative medium (see \fref{exp_setup}). In particular, feedback consists of the display of arrows on a screen, representing the direction and magnitude of the correct action $\optimalactions$ and the action provided by the user $\bu$. Additionally, a progress bar at the top tracks the participant's teaching effort throughout the task, advancing after each episode. The position of the robot’s end effector is displayed along with additional arrows representing other states, such as velocity. These states are shown sequentially, with one state appearing after another, to ensure clarity and focus for the participant. This design is general and flexible, making it applicable to a wide range of robotic learning tasks and adaptable to different teaching environments.

% \begin{figure}[t]
%     \centering
%     \begin{subfigure}[b]{0.49\linewidth}
%         \centering
%         \includegraphics[width=\linewidth]{img/exp1_ui.png}
%         \caption{}\label{f:exp_setup_1}
%     \end{subfigure}
%     \hfill
%     \begin{subfigure}[b]{0.49\linewidth}
%         \centering
%         \includegraphics[width=\linewidth]{img/exp2_setup.png}
%         \caption{}\label{f:exp_setup_2}
%     \end{subfigure}
%     \hfill
%     % \begin{subfigure}[b]{0.49\linewidth}
%     %     \centering
%     %     \includegraphics[width=\linewidth]{img/exp2_setup1.png}
%     %     \caption{}\label{f:exp_setup_3}
%     % \end{subfigure}
%     % \hfill
%     \begin{subfigure}[b]{0.49\linewidth}
%         \centering
%         \includegraphics[width=\linewidth]{img/exp2_setup2.png}
%         \caption{}\label{f:exp_setup_4}
%     \end{subfigure}
%     \caption{Experiment Setup}
%     \label{f:Exp1Result1}
% \end{figure}

\begin{figure}[t]
    \centering
    \includegraphics[width=\linewidth]{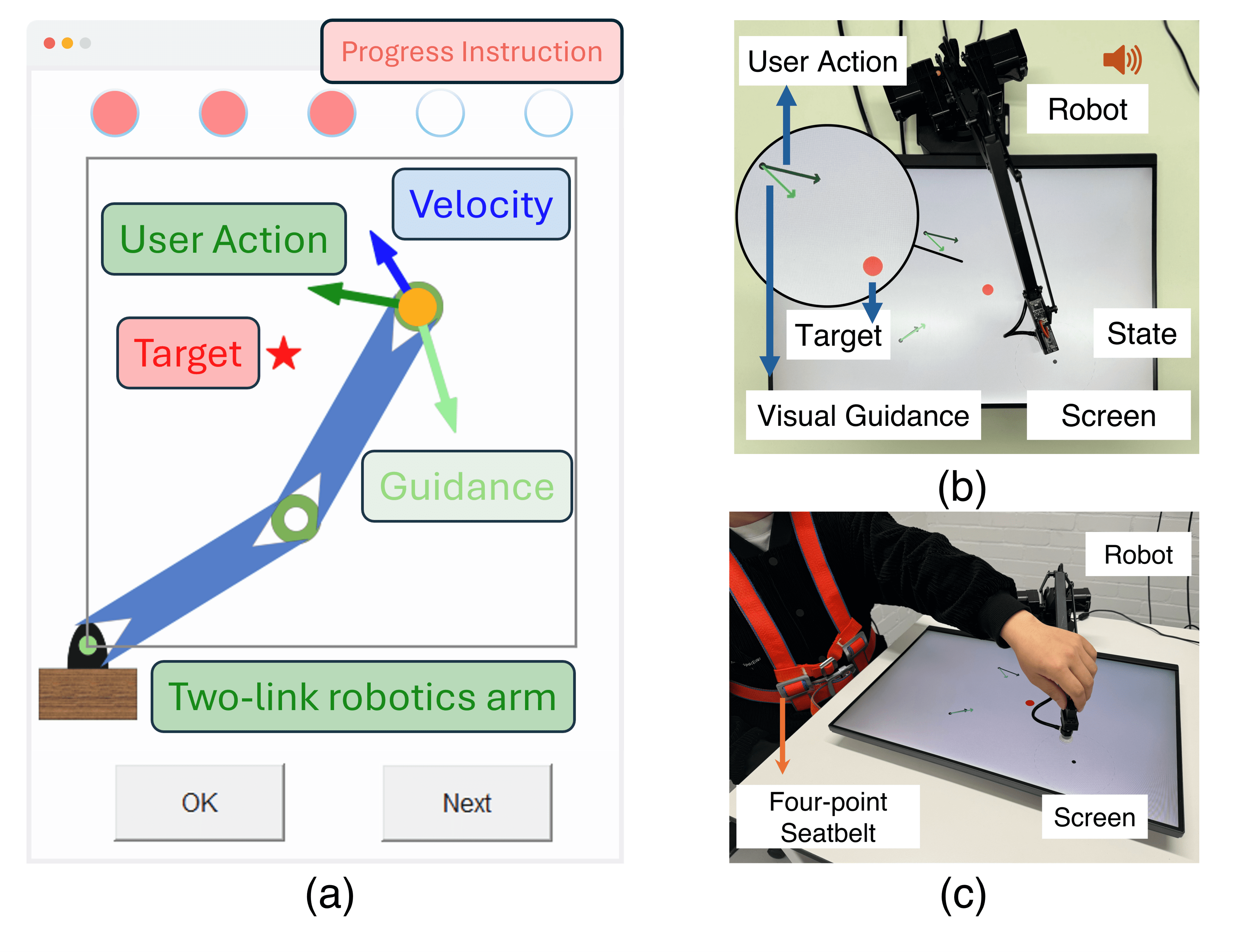}
    \caption{(a) Simulation Platform User Interface: The interface displays a two-axis robotic arm fixed at the origin. The yellow marker and the blue arrow represent the robot's position and velocity, respectively. The dark green arrow shows the action applied by the user, and the light green arrows represent the \MT-based guidance. The progress bar shows the teaching effort so far. (b) Physical Experiment Setup (Top View): The uArm includes a display screen for visual guidance. The black dot shows the current state, and the red dot marks the target. The light green arrow shows the user’s action, and the dark green arrow represents \MT-based guidance. Auditory guidance for teaching effort progress is provided via a speaker. (c) Physical Experiment with Participant: A four-point seat belt ensures the participant maintains a consistent viewpoint.}
    \label{f:exp_setup}
\end{figure}

\section{Evaluation}
\label{s:evaluation}
In this section, two experiments\footnote{This study is conducted under the approval of the King's College London Research Ethics Committee, Ref.: MRPP-22/23-37844. Informed consent was obtained from all experimental participants. The data collected for this research is open access, with accreditation, from {\tt\small http://doi.org/[link to be created upon acceptance]}.} are reported in which the effectiveness of using this training framework to train novice human teachers is assessed in \il{\item a simulated platform and \item a physical robot}.

\subsection{Experimental Protocol} \label{s:protocol}
% The following outlines a between-subject experiment in which participants are randomly assigned to a target or a control group, and asked to provide teaching demonstrations in a sequence of phases:
% \begin{enumerate}[label=P\arabic*]
%     \item\label{i:phase1} \textit{Skill 1, no guidance.} Participants give demonstrations to teach $\ref{i:s1}$ without guidance (1 episode).
%     \item\label{i:phase2} \textit{Skill 2, no guidance.} Participants give demonstrations to teach $\ref{i:s1}$ without guidance (1 episode).
%     \item\label{i:phase3} \textit{Skill 1, with guidance.} Participants give demonstrations to teach $\ref{i:s1}$. Those in the target group are given guidance, while those in the control group are not (8 episodes).
%     \item\label{i:phase4} Participants repeat \ref{i:phase1} (1 episode).
%     \item\label{i:phase5} Participants repeat \ref{i:phase2} (1 episode).
% \end{enumerate}

The following outlines a between-subject experiment in which participants are randomly assigned to a target or control group and asked to provide teaching demonstrations in a sequence of phases (P1-P5), with each phase comprising a number of episodes (E):
% For instance, P3E1 refers to Phase 3, Episode 1. The detailed protocol is as follows:

\begin{enumerate}[label=P\arabic*]
    \item\label{i:phase1} \textit{Skill 1, no guidance.} Participants give demonstrations to teach $\ref{i:s1}$ without guidance \textbf{(P1E1)}.
    \item\label{i:phase2} \textit{Skill 2, no guidance.} Participants give demonstrations to teach $\ref{i:s2}$ without guidance \textbf{(P2E1)}.
    \item\label{i:phase3} \textit{Skill 1, with guidance.} Participants give demonstrations to teach $\ref{i:s1}$. Those in the target group are given guidance, while those in the control group are not \textbf{(P3E1-E8)}.
    \item\label{i:phase4} Participants repeat \ref{i:phase1} \textbf{(P4E1)}.
    \item\label{i:phase5} Participants repeat \ref{i:phase2} \textbf{(P4E1)}.
\end{enumerate}

The protocol are designed to first establish the participants' baseline performance of teaching ability in two motor skills (\ref{i:phase1}, \ref{i:phase2}). Then in the treatment phase (\ref{i:phase3}), participant in two groups are trained under different conditions across eight episodes to evaluate the impact of the guidance. The \ref{i:phase4} is designed to assess post-treatment retention of teaching ability after training while \ref{i:phase5} is designed to evaluate the transfer of teaching ability to a new skill.
% The phases are designed to first assess participants' baseline teaching ability (\ref{i:phase1}, \ref{i:phase2}), followed by an evaluation of immediate guided training's impact (\ref{i:phase3}). \ref{i:phase4} assesses whether the improvements are retained without guidance, while \ref{i:phase5} tests the generalisation of improved teaching to a new task.
Each phase of teaching may consist of one or more episodes, in which a participant gives a complete set of demonstrations on a set of pre-determined query states. Note that, the query states differ in each phase of teaching.

In \ref{i:phase3}, the \emph{target group} receives real-time visual guidance, displaying both demonstrated and optimal actions (see \fref{exp_setup}), while the \emph{control group} proceeds without guidance. After each phase, all participants receive feedback in the form of skill reproduction (\ie the robot plays back the learnt behaviour). Prior to the experiment, participants received video\footnote{The instruction video used in this experiment is available to view online at https://youtu.be/Ir2DLJAYytY.} instructions and two minutes of unstructured practice time. No time limits were applied during the teaching phases.
% \footnote{The introductory video used in the experiment is provided as a supplementary file to this paper, and is available to view online}

% Based on G*Power calculations (effect size = 0.9, error probability = 0.05, power = 0.8), a total of $\nsubjects=32$ participants were recruited for each experiment and randomly assigned to either the target group or the control group.
% Sample size was determined using G*Power, with an effect size of 0.9, an error probability of 0.05, and a power of 0.8, resulting in a required sample size of $\nsubjects=32$ participants for each experiment. A single-blind procedure was implemented to ensure participants were unaware of their group assignment. Eligibility criteria required that participants had no professional experience or background in robotics or related fields.

The sample size ($\nsubjects=32$ for each experiment) was calculated in \emph{GPower 3.1} \cite{faul2009statistical} using a one-tailed t-test, with an effect size of $0.9$, an alpha level of $0.05$, and a statistical power of $0.8$. A single-blind procedure was used to ensure they were unaware of their group assignment. Eligibility criteria required that participants had no professional experience or background in robotics.

%\subsubsection{Evaluation Metrics}
The teaching quality of the human teacher is directly reflected in the robot's learning performance. In this study, since the $\target$ is predefined, teaching quality is evaluated by the difference between the target and the learned controller parameters%, using the \eref{l2norm}:
% Teaching quality is measured by the difference between the target and learned controller parameters, using the \eref{l2norm}:
\begin{equation} \label{e:l2norm}
    \ltwonorm = \sqrt{(\learnt-\target)^\T(\learnt-\target)},\quad\target=\vect(\Model).
\end{equation}
% where $\theta = vec(\Theta)$.
This metric provides a benchmark for how closely the robot's learned controller matches the target. But note, this is available only in this controlled experiment, as $\target$ is often unknown in some scenarios.

\subsection{Simulated Robot}\label{s:simulation_exp}%
The aim of the first evaluation is to test the proposed framework in a highly-controlled environment. To achieve this, a physical simulation of a two-link robot arm is used to assess novice users' ability to teach force-controlled tasks. 
\subsubsection{Set up}
The robot's state is represented as $\state=(\qi,\qii,\qdoti,\qdotii)^\T$, where $\bq=(\qi, \qii)^\T$ and $\bqdot=(\qdoti, \qdotii)^\T$ %$\br=(\xi, \yi)^\T$ and $\brdot=(\xii, \yii)^\T$ 
denote the position and velocity of the robot's joints, respectively. 
The actions consist of the force in end-effector space $\action = (\fx, \fy)^\T$. Each link of the arm has length $\length=\SI{1}{\meter}$ and mass $\mass=\SI{1}{\kilogram}$. Gravity $\gravity$ is considered to be \SI{9.81}{\meter\per\second\squared}. No force limits are apply to the arm, and it operates at a sampling rate of \SI{1}{\kilo\hertz}. 

The dynamics of the robot arm are modelled using the standard mass, coriolis and gravity formulation:
\begin{equation}
    \label{e:MCG}
    \bM(\bq) \bqddot + \bC(\bq, \bqdot) \bqdot + \bg(\bq) = \bJ^\T(\bq) \action
\end{equation}
where $\bqddot$ is the joint angluar acceleration, $\bM\in\R^{2\times2}$ is the mass matrix, $\bC\in\R^{2\times2}$ represents Coriolis and centrifugal forces and $\bg\in\R^{2\times1}$ is the gravity vector. The applied $\action$ is converted into joint torques via the Jacobian $\bJ\in\R^{2\times2}$, ensuring accurate force control at the end-effector.

For the teaching process, an online interactive tool is used, as shown in \fref{exp_setup}(a). The tool generates a set of query states according to \sref{selection_of_states}, and the teacher demonstrates the appropriate $\action$ by applying directional forces by drawing on the interface. After each demonstration, \MT-based visual guidance, in the form of arrows, may be shown as guidance. After all demonstrations in an episode have been given, the robot reproduces the learned skill from a fixed starting point. This is shown as an animation to the teacher. The availability of these feedback mechanisms (visual guidance and skill reproduction) is controlled by the experimental condition (see \sref{protocol}).

The class of skills that must be taught to the robot consists of \emph{convergent motion} from any point in the workspace to a target. Specifically, the two skills studied here are to reach to
\begin{enumerate*}[label=$\Model_{\arabic*}$,ref=\Model_{\arabic*}]
    \item\label{i:s1} a \emph{target point}, and
	\item\label{i:s2} a \emph{target line},
\end{enumerate*}
which are consistent with the task of a robotic arm moving goods to wooden pallets or conveyor belts in a factory.
These skills can be represented by a coupled pair of controllers of the form \eref{LiPSkillModel} with $\bphi(\state)=(\xi,\yi,\xii,\yii,1)$ where $\br=(\xi, \yi)^\T$ and $\brdot=(\xii, \yii)^\T$ 
denote the position and velocity of the robot's end-effector, respectively, and
\begin{equation}
    \Model_i = \left(\begin{array}{ccccc}
k_1 & k_2 & d_1 & d_2 & \targetx\\ k_3 & k_4 & d_3 & d_4 & \targety
\end{array}
\right),\quad i\in\{1,2\}
\end{equation}
%with $\model=\vect(\Model)$.
%
with %In the below, 
$\vect{(\ref{i:s1})}=(-1,0,0,-1,-1,0,0,-1,0.8,1.2)^\T$ and $\vect{(\ref{i:s2})}=(-1,0,0,0,-1,0,0,-1,0.8,0)^\T$. 
%
%In the below, $\ref{i:s1}=(\model_{1}^{1}, \model_{1}^{2})^\T$: $\model_{1}^{1} = (-1,0,-1,0,0.8)^\T$, $\model_{1}^{2} = (0,-1,0,-1,1.2)^\T$; $\ref{i:s2}=(\model_{2}^{1}, \model_{2}^{2})^\T$: $\model_{2}^{1} = (-1,0,-1,0,0.8)^\T$, $\model_{2}^{2} = (0,-1,0,0,0)^\T$.
%
%\begin{equation}
%    \Model = (\model^x, \model^y)^T = \left(\begin{array}{ccccc}
%k_1 & k_2 & d_1 & d_2 & \targetx\\ k_3 & k_4 & d_3 & d_4 & \targety
%\end{array}
%\right)
%\end{equation}
%
%\begin{equation}
%    \Model = (\model^x, \model^y)^T = \left(\begin{array}{ccccc}
%k_1 & k_2 & d_1 & d_2 & \targetx\\ k_3 & k_4 & d_3 & d_4 & \targety
%\end{array}
%\right)
%\end{equation}
%
%Herein, $\ref{i:s1}=(\model_{1}^{x}, \model_{1}^{y})^\T$: $\model_{1}^{x} = (-1,0,-1,0,0.8)^\T$, $\model_{1}^{y} = (0,-1,0,-1,1.2)^\T$; $\ref{i:s2}=(\model_{2}^{x}, \model_{2}^{y})^\T$: $\model_{2}^{x} = (-1,0,-1,0,0.8)^\T$, $\model_{2}^{y} = (0,-1,0,0,0)^\T$.
%
% \begin{equation}
%     \Model = \left(\begin{array}{ccccc}
% \model^x \\ \model^y
% \end{array}
% \right) = \left(\begin{array}{ccccc}
% k_1 & k_2 & d_1 & d_2 & \targetx\\ k_3 & k_4 & d_3 & d_4 & \targety
% \end{array}
% \right)
% \end{equation}

The results are based on a sample of $\nsubjects=32$ participants, consisting of an equal proportion of males and females, with an average age of $27.9$ years (SD = $7.7$).

% \begin{figure}[t!]
%     \centering
%     \includegraphics[width=0.5\textwidth]{img/exp1.png}
%     \caption{Simulation Experiment Results. \subref{f:exp1_results} indicates the change in the mean $\ltwonorm$ for the two groups as the experiment progresses. \subref{f:exp1_result_p1p3}-\subref{f:exp1_result_p2p5} show the change in $\ltwonorm$ between \ref{i:phase1} and \ref{i:phase3}E8, \ref{i:phase1} and \ref{i:phase4}, \ref{i:phase2} and \ref{i:phase5}, respectively.}\label{f:exp1_result_1}
% \end{figure}

\begin{figure}[t!]
    \centering
    \includegraphics[width=0.5\textwidth]{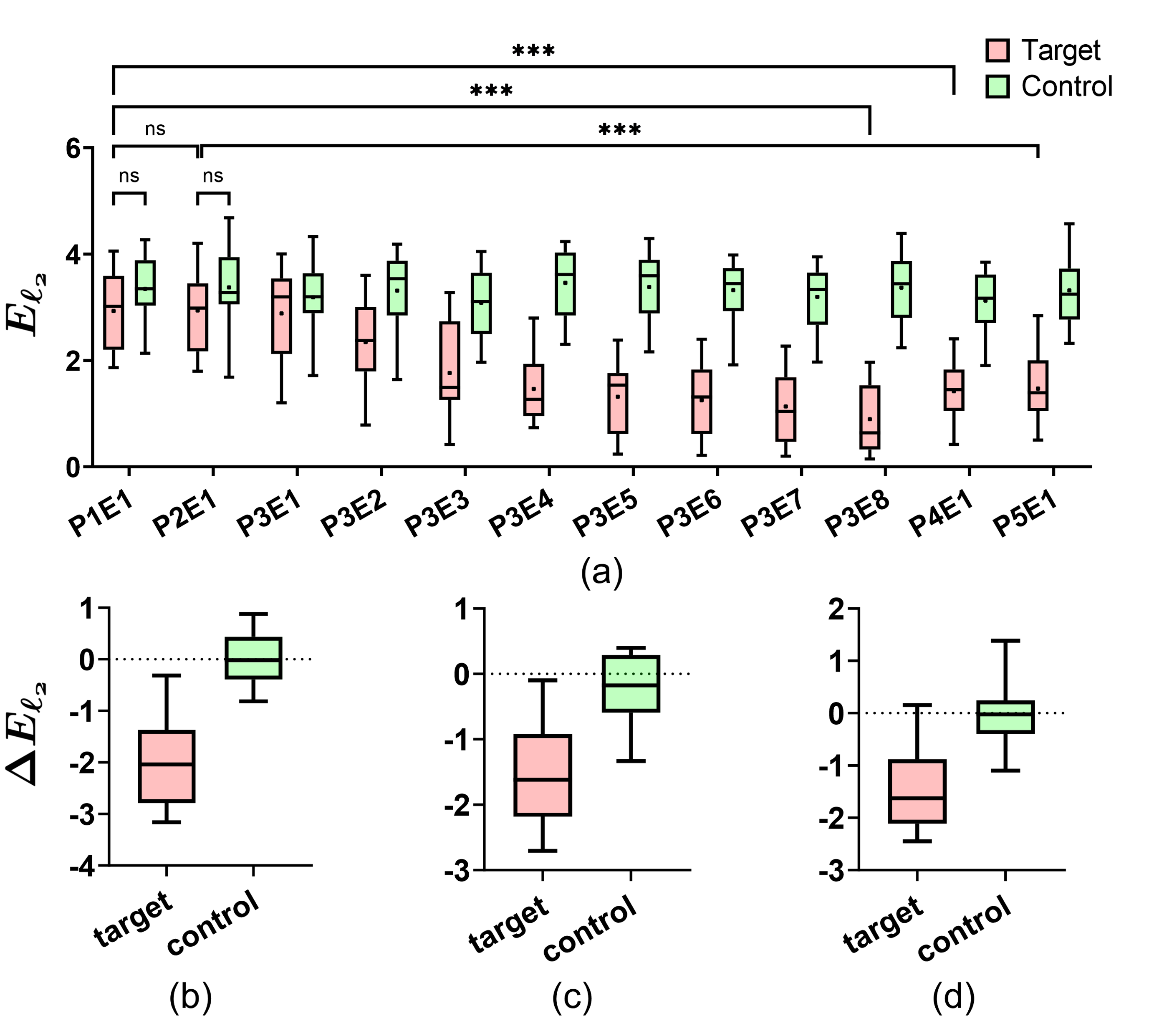}
    \caption{Simulation Experiment Results. (a) indicates the change in the mean $\ltwonorm$ for the two groups as the experiment progresses. (b)-(d) show the change in $\ltwonorm$ between \ref{i:phase1} and \ref{i:phase3}E8, \ref{i:phase1} and \ref{i:phase4}, \ref{i:phase2} and \ref{i:phase5}, respectively.}\label{f:exp1_result_1}
\end{figure}

\subsubsection{Results} The results are summarised in \fref{exp1_result_1}. A post-hoc analysis of data between \ref{i:phase1} and \ref{i:phase2} showed no significant difference in teaching behaviour between the two skills and two groups ($p = 0.0989$, $p = 0.1148$).

% \paragraph{Effectiveness of Guidance} \fref{exp1_result_1}(a) shows the trend in $\ltwonorm$ for the target and control groups throughout \ref{i:phase3}. The target group, which received real-time visual guidance, shows a steady reduction in median error as the phase progresses, with an overall improvement of approximately 63\%. In contrast, the control group, which did not receive \MT-based guidance, shows no noticeable downward trend, with their median error remaining stable. Without structured feedback, participants in the control group struggled to improve, relying on intuition, which did not lead to substantial gains in teaching accuracy. Statistical analysis supports these observations, as the error reduction for the target group between \ref{i:phase1}E1 and \ref{i:phase3}E8 is significant ($p<0.05$), while no significant change was observed in the control group ($p>0.05$) (\fref{exp1_result_1}(b)). These results show \ref{h1} is supported.%, demonstrating that \MT-based guidance significantly enhances teaching quality.

\paragraph{Effectiveness of Guidance} \fref{exp1_result_1}(a) shows the trend in teaching error for the target and control groups throughout all phases. During \ref{i:phase3}, the target group, which received real-time visual guidance, shows a steady reduction in mean error as the episodes progresses, with an overall improvement of approximately 69.40\%. In contrast, the control group, which did not receive \MT-based guidance, shows no noticeable downward trend, with their mean error remaining stable. In addition, \fref{exp1_result_1}(b) highlights the change in $\ltwonorm$ between the treatment (\ref{i:phase3}E8) and the baseline (\ref{i:phase1}E1) phase. The target group ($-2.03 \pm 0.82$) showed a greater reduction compared to the control group ($0.025 \pm 0.53$). Without structured feedback, participants in the control group struggled to improve, relying on intuition, which did not lead to substantial gains in teaching accuracy. 
% Statistical analysis supports these findings, with the target group showing a significant reduction in teaching error between \ref{i:phase1}E1 and \ref{i:phase3}E8 ($p<0.001$), while no significant change was observed in the control group ($p>0.05$) . These results confirm that \ref{h1} is supported.
These results confirm that \ref{h1} is supported.
%, demonstrating that \MT-based guidance significantly enhances teaching quality.

\paragraph{Retention of Teaching Quality} 
% \fref{exp1_result_1}(c) compares \ref{i:phase1}E1 and \ref{i:phase4}E1, which involve the same states before and after the guidance phase. The error in \ref{i:phase4}E1 drops by approximately 50\% compared to \ref{i:phase1}E1, indicating that participants retained the improvements made through \MT-based guidance. Although the median error in \ref{i:phase4}E1 is slightly higher than in \ref{i:phase3}E8 (see \fref{exp1_result_1}), it still shows that the participants' teaching strategies learned during the guided phase are carried forward, benefiting them even in the absence of guidance.
% The findings provide evidence for \ref{h2}.

\fref{exp1_result_1}(c) compares the change in teaching error between \ref{i:phase1} and \ref{i:phase4} for the target group ($-1.51 \pm 0.84$) and the control group ($-0.22 \pm 0.56$). The target group's mean error was 51.47\% lower in \ref{i:phase4} compared to \ref{i:phase1}, indicating that participants retained the improvements made through \MT-based guidance. Although the mean error in \ref{i:phase4} is slightly higher than in \ref{i:phase3}E8 (see Fig 3), this still shows that the participants' teaching strategies learned during the guided phase are carried forward, benefiting them even in the absence of guidance. These findings supports \ref{h2}.

% \fref{exp1_result_p1p4} shows the comparison of \ref{i:phase1}E1 and \ref{i:phase4}E1 between both group.

\paragraph{Generalisation to New Skills} 
% \fref{exp1_result_1}(d) compares the performance between \ref{i:phase2}E1 and \ref{i:phase5}E1, evaluating whether the improvements from \MT-based guidance generalise to teaching a new skill. After receiving guidance in \ref{i:phase3}, the target group demonstrate a notable improvement in \ref{i:phase5}E1, with the error dropping by approximately 56\% compared to \ref{i:phase2}E1, whereas the control group show no change in performance. This shows that the teaching strategies learned for the first skill are successfully transferred to teaching the new task, demonstrating the ability of \MT-based guidance to help participants generalise their improvements across different contexts. The data validates \ref{h3}.

\fref{exp1_result_1}(d) compares the performance between \ref{i:phase2} and \ref{i:phase5}, evaluating whether the improvements from \MT-based guidance generalise to teaching a new skill. It highlights the reduction in teaching error between \ref{i:phase2} and \ref{i:phase5} for the target group ($-1.47 \pm 0.80$) and the control group ($-0.06 \pm 0.64$). After receiving guidance, the target group demonstrate a notable improvement in \ref{i:phase5}, with the error dropping by approximately 50.03\% compared to \ref{i:phase2}, whereas the control group show no change in performance. This shows that the teaching strategies learned for the trained skill are successfully transferred to teaching the new task, demonstrating the ability of \MT-based guidance to help participants generalise their improvements across different contexts. The data supports \ref{h3}.

The results from the simulation experiment clearly demonstrate the effectiveness of the proposed \MT-based guidance mechanism in improving teaching accuracy, retaining learned strategies, and generalising them to new tasks. This confirms the value of structured feedback in enhancing both short-term and long-term teaching quality for novice teachers.

\subsection{Physical Robot Experiment}
The aim of the second experiment is to assess whether the proposed training approach is effective when teaching physical robotic systems, in face of the inherent mechanical variability, sensor noise, and hardware constraints, found in real-world environments.

\subsubsection{Set Up}
The robotic platform chosen for this experiment is the uArm Swift Pro. This is a 4-degree-of-freedom robotic arm with 
% $\length=\SI{1}{\meter}$ and mass $\mass=\SI{1}{\kilogram}$. Gravity $\gravity$ is considered to be \SI{9.81}{\meter\per\second\squared}.
two \SI{0.15}{\meter} links and a total weight of \SI{2.2}{\kilogram}. It operates within a working range of \SI{50}{\milli\meter} to \SI{320}{\milli\meter} and reaches a maximum speed of \SI{100}{\milli\meter\per\second}. With stepper motors and a 12-bit encoder, it achieves a positional repeatability of \SI{0.2}{\milli\meter}, ensuring high precision in training tasks. The system operates with a maximum torque limit of \SI{12}{\kilogram\centi\meter}. More detailed technical specifications can be found in the manufacturer's documentation \cite{uArm_doc}.

The experimental set up is shown in \fref{exp_setup}(b)-(c). In this experiment, the robot is limited to work in the horizontal plane by restricting the range of the joints. This enables real-time feedback to be displayed during the \LfD\ process via a \SI{\screenx}{\milli\meter} $\times$ \SI{\screeny}{\milli\meter} display. 

%The suction cups on the end-effector simulate the common \emph{convergent motion} used in industry, particularly for handling objects to a specified position or onto a designated conveyor belt.

The robot is kinematically controlled, with its state defined by the instantaneous position of its joints $\bq=(\qi,\qii,\q_3,\q_4)^\T$ and the %position of its end-effector: $\state=({\xi},{\yi})^\T$, where ${\xi}$ and ${\yi}$ denote the instantaneous position of the robot's end-effector. 
%The corresponding feature vector is $\phi(\state)=[{\xi},{\yi},1]^\T$, and the 
action given by $\action = (\udx, \udy)^\T$, where $\udx$ and $\udy$ represent the desired change in the end-effector position. %Velocity is omitted to simplify the task for participants, avoiding cognitive overload. 
Note that, the uArm Swift Pro, like many industrial robots %(e.g., Universal Robots UR3, ABB’s YuMi), 
is designed to prioritise positional accuracy over velocity or force. %This aligns with many real-world applications, such as assembly and material handling, where positional control is the primary focus. Simplifying the state-action space to position keeps the experiment practical without sacrificing complexity.

Similar to the preceding experiment, the motor skills to be taught represent \emph{convergent motion to a target} defined by a pair of coupled controllers with parameters %$\model=\vect(\Model)$ where
\begin{equation}
    \label{e:states}
    \Model_i = \left(\begin{array}{ccc}
            k_1 & k_2 & \targetx\\ k_3 & k_4 & \targety
            \end{array}
            \right),\quad i\in\{1,1\}
\end{equation}

with the corresponding feature vector $\bphi(\state)=({\xi},{\yi},1)^\T$ where ${\xi}$ and ${\yi}$ denote the instantaneous position of the robot's end-effector. This aligns with many real-world applications, such as assembly and material handling, where positional control is the primary focus. Specifically, here, $\vect{(\ref{i:s1})}=(-0.2,0,0,-0.2,4.6,2.2)^\T$ and $\vect{(\ref{i:s2})}=(-0.04,0.08,0.08,-0.16,0,0)^\T$. With the exception of the aforementioned differences in the robotic platform and skill representation, the experiment otherwise follows the protocol described in \sref{protocol}. The following reports results from $\nsubjects=32$ (new) participants (18 males and 14 females, of age $26 \pm7.3$).

% \begin{figure}[t!]
%     \centering
%     \begin{minipage}[b]{0.5\textwidth}
%         \centering
%         \includegraphics[width=\textwidth]{img/exp2_1.png}
%         \subcaption{}\label{f:exp2_results}
%     \end{minipage}    
%     \begin{minipage}[b]{0.15\textwidth}
%         \centering
%         \includegraphics[width=\textwidth]{img/exp2_2-1.png}
%         \subcaption{}\label{f:exp2_result_p1p3}
%     \end{minipage}
%     \hfill
%     \begin{minipage}[b]{0.15\textwidth}
%         \centering
%         \includegraphics[width=\textwidth]{img/exp2_2-2.png}
%         \subcaption{}\label{f:exp2_result_p1p4}
%     \end{minipage}
%     \hfill
%     \begin{minipage}[b]{0.15\textwidth}
%         \centering
%         \includegraphics[width=\textwidth]{img/exp2_2-3.png}
%         \subcaption{}\label{f:exp2_result_p2p5}
%     \end{minipage}
%     \caption{Physical Experiment Results. \subref{f:exp1_results} indicates the change in the mean $\ltwonorm$ for two groups as the experiment progresses. \subref{f:exp2_result_p1p3}-\subref{f:exp2_result_p2p5} show the changes of $\ltwonorm$ between: \ref{i:phase1} and \ref{i:phase3}, \ref{i:phase1} and \ref{i:phase4}, \ref{i:phase2} and \ref{i:phase5}, respectively.}\label{f:exp2}%
% \end{figure}

\begin{figure*}[t]
    \centering
    \begin{minipage}[b]{\textwidth}
        \centering
        \includegraphics[width=\textwidth]{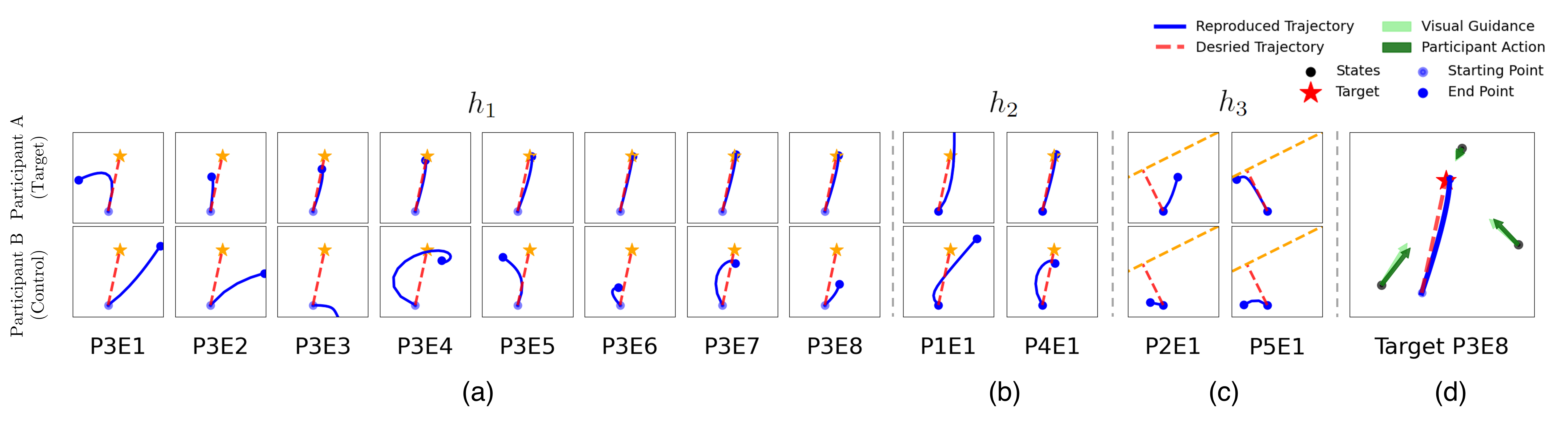}
        % \subcaption{}
        % \label{}
    \end{minipage}
    \caption{Trajectory Comparison. Trajectories produced by representative participants A (from the target group) and B (from the control group). (a) Trajectories from \ref{i:phase3}; (b) Trajectories from \ref{i:phase1} and \ref{i:phase4}; (c) Trajectories from \ref{i:phase2} and \ref{i:phase5}; (d) Zoomed-in view of participant A's trajectory from \ref{i:phase3}.}\label{f:exp2_result_3}
\end{figure*}

\begin{figure}[t!]
    \centering
    \includegraphics[width=0.5\textwidth]{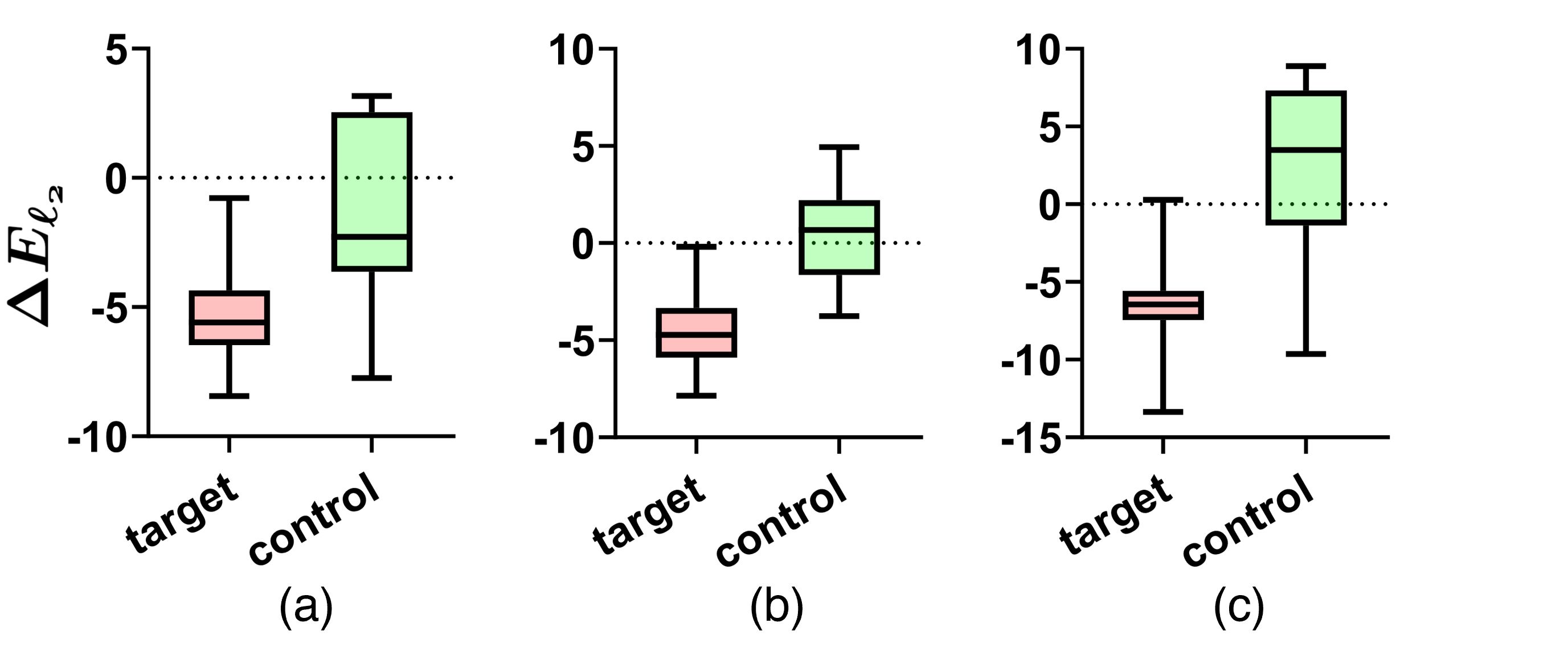}
    \caption{Simulation Experiment Results. (a)-(c) show the change in $\ltwonorm$ between \ref{i:phase1}E1 and \ref{i:phase3}E8, \ref{i:phase1}E1 and \ref{i:phase4}E1, \ref{i:phase2}E1 and \ref{i:phase5}E1, respectively.}\label{f:exp2_result_2}
\end{figure}

\subsubsection{Results} The results are summarised in \fref{exp2_result_2}. Post-hoc analysis of the data comparing \ref{i:phase1} and \ref{i:phase3} showed no significant variation in teaching behaviour between both skills and groups ($p > 0.05$).

\paragraph{Effectiveness of Guidance} During \ref{i:phase3} the teaching error in the target group still shows a continuous downward trend, with a 88.26\% reduction in \ref{i:phase3}E8 compared to \ref{i:phase1}. In contrast, without the guidance, the control group's teaching error fluctuates throughout \ref{i:phase3}. This is reflected in \fref{exp2_result_2}(a), where the drop in teaching error between \ref{i:phase1} and \ref{i:phase3} in the target group ($-5.30\pm 1.79$) is greater than that of the control group ($-1.54 \pm 3.44$), supporting \ref{h1} that examine the effectiveness of guidance.

\paragraph{Retention of Teaching Quality} \fref{exp2_result_2}(b) shows that the improvement in teaching quality from \ref{i:phase1} to \ref{i:phase4} is greater in the target group ($4.55 \pm 1.80$) compared to the control group ($-0.35 \pm 2.46$). This result further supports \ref{h2}, i.e. the retention of teaching ability achieved through our proposed framework.

\paragraph{Generalisation to New Skills} 
Looking at \fref{exp2_result_2}(c), the change in teaching error in the target group ($ -6.32 \pm 2.92$), which has a 77.34\% reduction from \ref{i:phase2} to \ref{i:phase5}, is significantly greater in the control group ($2.21 \pm 5.54$).
% The teaching error for the target group between \ref{i:phase2} and \ref{i:phase5} has a 77.34\% reduction, while the control group's error actually increases by 40\%. These changes are also reflected in \fref{exp2_result_2}(c). 
The observation presents further evidence supporting \ref{h3}, showing that the teaching ability developed through the proposed framework is effectively transferred across skills.

\paragraph{Teaching Strategy}
\fref{exp2_result_3} presents the trajectories reproduced by the robot, which were learnt from demonstrations provided by two representative participants in both groups. In \ref{i:phase3}, the target group demonstrates an increasing alignment with the desired trajectory. In contrast, the control group attempts to correct their initial demonstration but failed. Their trajectories in \ref{i:phase4} and \ref{i:phase5} remain inconsistent and distant from the target. When comparing \ref{i:phase4} and \ref{i:phase5}, it shows that the target group effectively retained their improved ability and could generalise it to new query states, and even to a new motor skill. Conversely, the control group struggled to achieve these retention and generalisation.

When observing how the control group provided actions, two key observations emerge. First, in the initial few phases, users often provide actions that point to the target, with a large magnitude for both groups. This pattern is consistent with the findings in the active learning literature, where participants tend to use ‘extreme’ examples to convey their teaching strategies \cite{khan2011humans}. Second, during continuous exploration, most control group users only explore the direction of action and failed to adjust the magnitude of their inputs, resulting in ineffective strategy adjustments. In contrast, the target group became more cautious and deliberate in \ref{i:phase3}. Their actions increasingly reflected a balanced consideration of both magnitude and direction, leading to improved teaching performance in \ref{i:phase4} and \ref{i:phase5}, as shown in \fref{exp2_result_3}(d).

\section{CONCLUSIONS}\label{s:conclusion}%
This study proposes an \MT-based guidance framework designed to assist novice teachers in providing high-quality demonstrations of dynamic skills to robots. In experiments in \LfD\ of simulated and learnt robots, use of the proposed framework enables users to improve their teaching ability, and this is both \il{\item retained over time, and \item generalises across skills}. %effectiveness was validated through experiments on both simulation and real robots, with quantitative analysis of teaching quality and qualitative analysis of user strategies. 
The results show that the framework improves teaching quality by 78.83\% for trained skills and by 63.69\% for generalised skills. Future work will focus on evaluating the framework's effectiveness on more complex industrial robots, % supporting workers in robot manipulation through \LfD\, and 
including testing it in a real industrial production line environment.

% \newpage
% \clearpage
{\scriptsize\printbibliography}
% \addtolength{\textheight}{-5cm}   % This command serves to balance the column lengths
\cleardoublepage
%\section*{APPENDIX}
%\section*{ACKNOWLEDGMENT}
%%%%%%%%%%%%%%%%%%%%%%%%%%%%%%%%%%%%%%%%%%%%%%%%%%%%%%%%%%%%%%%%%%%%%%%%%%%%%%%%
% {\scriptsize\printbibliography}%
\end{document}